\documentclass[11pt]{article}
\usepackage[utf8]{inputenc}
\usepackage[T1]{fontenc}
\usepackage{coling2018}
\usepackage{times}
\usepackage{url}
\usepackage{latexsym}
\usepackage{array}
\usepackage{xspace}
\usepackage{multirow}
\usepackage{graphicx}
\usepackage[utf8]{inputenc}
\usepackage{CJKutf8}

\newcolumntype{L}[1]{>{\raggedright\let\newline\\\arraybackslash\hspace{0pt}}m{#1}}
\newcolumntype{C}[1]{>{\centering\let\newline\\\arraybackslash\hspace{0pt}}m{#1}}
\newcolumntype{R}[1]{>{\raggedleft\let\newline\\\arraybackslash\hspace{0pt}}m{#1}}

\def\eg{\emph{e.g.}\xspace}			

\title{Design Challenges in Named Entity Transliteration}

\author{Yuval Merhav\thanks{*Y. Merhav and S. Ash contributed equally to this work} \\
  Amazon Alexa AI \\
  Cambridge, MA \\
  {\tt merhavy@amazon.com} \\\And
  Stephen Ash\footnotemark[1] \\
  Amazon AWS AI\\
  Seattle, WA \\
  {\tt ashstep@amazon.com} \\}

\date{}

\begin{document}

%
%
\blfootnote{
    %
    %
    %
    %
    
    
    \hspace{-0.65cm}  
    This work is licensed under a Creative Commons 
    Attribution 4.0 International License.
    License details:
    \url{http://creativecommons.org/licenses/by/4.0/}
}

\maketitle
\begin{abstract}
  We analyze some of the fundamental design challenges that impact the development of a multilingual state-of-the-art named entity transliteration system, including curating bilingual named entity datasets and evaluation of multiple transliteration methods. We empirically evaluate the transliteration task using the traditional weighted finite state transducer (WFST) approach against two neural approaches: the encoder-decoder recurrent neural network method and the recent, non-sequential Transformer method. In order to improve availability of bilingual named entity transliteration datasets, we release personal name bilingual dictionaries mined from Wikidata for English to Russian, Hebrew, Arabic, and Japanese Katakana. Our code and dictionaries are publicly available\footnote{\url{https://github.com/steveash/NETransliteration-COLING2018}}.
\end{abstract}

\section{Introduction}
\label{intro}

Named entity transliteration is the process of converting a named entity from one language script to another, using the correct characters that represent the entity in the target language. It is an important component in many search and language understanding tasks, such as robust cross-language information retrieval (CLIR) and machine translation (MT), among others.

A possible simple transliteration approach is mapping every character (or sequence of characters) in the source language to its most common counterpart in the target language. However, spelling and pronunciation of many languages (e.g., English, Japanese) can be ambiguous and inconsistent \cite{fushimi1999consistency}. As a result, most transliteration systems are data driven and use context for disambiguation; \eg, \cite{yan2016applying,haizhou2004joint,ekbal2006modified}.

While transliteration has been a long studied problem, some important aspects received little attention. 
There is not clear guidance that addresses a number of common design considerations faced when building a robust multilingual transliteration system, such as data representation and the huge gap in results depending on the language pairs and transliteration direction~\cite{rosca2016sequence}. Like many other NLP fields recently, \textit{neural} transliteration systems have gained popularity. However, it is still unclear if neural systems consistently outperform traditional approaches and what architecture is ideal for this task. For example, in \cite{yan2016applying} the authors applied a stack of convolutional layers and simple recurrent layer on top for English to Chinese transliteration, which achieved competitive results but still below a phrase-based statistical machine translation system. Other works show that bidirectional long short-term memory (LSTM) neural networks and the encoder-decoder architecture~\cite{sutskever2014sequence} achieve comparable results with WFST based n-gram models that are considered state-of-the-art~\cite{rao2015grapheme,yao2015sequence}. However, these works only studied grapheme-to-phoneme (G2P) conversion from English to standard English pronunciation sets, such as the CMU pronunciation dictionary~\cite{weide1998cmu}.
Also, there is not empirical evidence comparing the neural encoder-decoder method with LSTM to the recently proposed Transformer~\cite{vaswani2017attention} neural method on the transliteration task. The Transformer method uses a simple neural network architecture based solely on attention mechanisms. That motivated us to learn if it can produce strong results on transliteration as it did on translation.    

The contributions in this paper are summarized as follows:
\begin{itemize}
    \item We enumerate and experimentally evaluate a number of design considerations important to building a named entity transliteration system such as: handling infrequent or unique name tokens in training and test data, building a bilingual training corpus from Wikidata\footnote{\url{http://wikidata.org}}, and top-1 versus top-$k$ result performance.
    \item We present empirical results comparing design choices across four different language/script pairs mined from Wikidata: English to Hebrew, English to Russian, English to Arabic, and English to Katakana. We are releasing each of the bilingual datasets and the particular train, development, and test splits to encourage future experimentation and benchmarking.
    \item We empirically evaluate the traditional Weighted Finite State Transducer (WFST) approach against two neural network-based approaches: the sequence-to-sequence encoder-decoder architecture and the recent Transformer approach based of self-attention.
\end{itemize}

The rest of this paper is organized as follows: first, we briefly describe the traditional and neural approaches to transliteration present in the literature; second, we describe data collection considerations and our multilingual datasets; lastly, we describe our experimental setup, present a number of empirical results, and provide guidance based on our experience and analysis.

\section{Transliteration Approaches}
\subsection{Traditional Approaches}
Early transliteration works utilized dictionaries and phonetic resources to learn probabilistic mapping rules between languages~\cite{knight1998machine,stalls1998translating}. Later works introduced more robust methods that do not require an intermediate phonemic mapping and can learn direct orthographical mapping between two languages given a bilingual dictionary~\cite{haizhou2004joint}. Traditionally, such works are modeled based on the common Grapheme-to-Phoneme (G2P) joint-sequence modeling technique. Given word-to-pronunciation examples, an initial alignment between corresponding grapheme (word) and phoneme (pronunciation) sequences is learned. Then, a language model is learned on the aligned tokens (see \cite{bisani2008joint} for a detailed description of joint-sequence models for G2P). The open source Phonetisaurus~\cite{novak2012wfst} has shown to achieve state-of-the-art scores on different G2P tasks~\cite{thu2016comparison}.
Phonetisaurus implements the common EM-driven sequence alignment algorithm, with a few constraints such as allowing only $m$-to-one and one-to-$m$ alignments. It trains an n-gram language model from the aligned pairs, which is converted into a Weighted Finite-State Transducer (WFST). Decoding can then be done by extracting the shortest path through the phoneme lattice created via composition with the input word. 

\subsection{Neural Approaches}\label{ref:neuralApproaches}
Recent work in Neural Machine Translation (NMT) has proposed a number of approaches to use neural networks in variable-length sequence-to-sequence tasks such as transliteration. The encoder-decoder architecture~\cite{sutskever2014sequence} is a recurrent neural network setup with two parts. An \textit{encoder} is fed input tokens one at a time and encodes them into a hidden state vector. At the end of the input sequence, an end-of-sentence token is fed to signify the end of the encoding phase. Next, the hidden state output of the encoder is fed into the \textit{decoder}. The decoder emits tokens and updated hidden states, which are recursively fed into itself, until there are no more output tokens to produce. An additional mechanism, \textit{attention}~\cite{bahdanauCB14}, allows the decoder to focus on different parts of the input sequence and capture long-range dependencies. More recently, the Transformer~\cite{vaswani2017attention} model was proposed, which avoids the need for sequential processing, relying only on self-attention. A benefit of this approach is there is no information bottleneck in the encoded hidden state vector as in the Encoder-Decoder approach. Additionally, because there is no longer a sequential recurrent network, model training can be better parallelized, decreasing model training time.

Previous work applied variations of the encoder-decoder approach to the name transliteration task~\cite{rosca2016sequence} and grapheme-to-phoneme transduction~\cite{rao2015grapheme}, suggesting strong results on both. \cite{rosca2016sequence} reports that on English to Japanese Katakana transliteration a unidirectional encoder-decoder with 2 hidden layers using gated recurrent unit (GRU) as the underlying neural cell type achieves the best result: a word error rate (WER) of $0.50$.

\section{Data}\label{sec:datasets}

The transliteration shared task, as part of the named entities workshop (NEWS)\footnote{\url{http://workshop.colips.org/news2018/}}, has been a continuous effort of benchmarking different transliteration approaches and systems across different languages. The workshop released multiple multilingual datasets over the years. However, datasets from previous years are not publicly available and their license is restrictive. That motivated us to create new multilingual datasets based on Wikidata that will be publicly available and free for all.  

Wikipedia has been widely used in transliteration works (e.g.,~\cite{irvine2010transliterating,rosca2016sequence,pasternack2009learning}). Wikidata is a central knowledge base in all languages for Wikipedia and other Wikimedia projects. Most Wikidata pages contain labels\footnote{A Wikidata label is the most common name that the item would be known by. It does not need to be unique, in that multiple items can have the same label.} in one or multiple languages. We automatically collected all ``en'' (English) labels where they pair with one of the following labels: ``ja'' (Japanese), ``he'' (Hebrew), ``ar'' (Arabic), and ``ru'' (Russian). We filtered out labels containing characters not belonging to their main script (e.g., English labels containing non-Latin characters). For Japanese, we only included labels with Katakana characters. For English tokens, we did \textit{not} strip out diacritics\footnote{We did not remove any Latin script characters} as these may carry useful information in the context of named entity transliteration. We did convert all characters to lowercase and strip some punctuation such as underscores, braces, exclamation marks, etc.

The first version of our dataset contained many types of entities, such as song and book titles. A quick analysis of the data revealed that many label pairs are translations and not transliterations. Consequently, we used the Wikidata ``instance-of'' property to only include labels of type ``human''. There are significantly more pages of this type than any other type on Wikidata, and they are less noisy than other types for the transliteration task. Since no direction specific information was used in data gathering, we evaluate performance on both directions (\eg English $\rightarrow$ Arabic and Arabic $\rightarrow$ English).

We also report performance on two publicly available datasets: (1) The CMU pronunciation dictionary~\cite{weide1998cmu}; and (2) The Arabic to English dataset extracted from Wikipedia titles in~\cite{rosca2016sequence}. The former is used mainly for evaluating G2P systems and not named entity transliteration, but the two tasks are related and many papers use it solely for evaluation. 

\section{Experimental Setup}

\subsection{Data Representation}
Pronunciation of English names usually does not carry context across tokens. In the majority of names, the pronunciation of a last name is independent of the previous tokens (middle or first names). Hence, it makes sense to train a transliteration system on name pairs that consist of a single token on each side. Our Wikidata datasets contain many multi-token names. Consequently, for constructing the train and test examples, we learn the word alignments of multi-token names, which is a simple task since the majority of name pairs contain the same word order, and we are not dealing with CJK languages with no token boundaries. The majority of our Katakana names contain a middledot to separate the tokens. We throw away English to Katakana name pairs if the English name contains more than one token and the Katakana name has no middledots or spaces. It is worth noting that the majority of Russian names in Wikidata are written as ``last, first'' and not ``first last'' as most other languages. 

After alignment, we construct our cross-validation splits with only single, unique name tokens. Splitting names into single token examples produces many duplicates. For example, every name that starts with ``John'' would produce a ``John'' example. For train, development, and test splits, we only include a single instance for each unique name token. When we encounter multiple possible target script transliterations for the same English token, we only include the most frequent target transliteration\footnote{We resolve ties by picking one of the targets arbitrarily}. We found that including multiple transliteration targets instead of only the single most-frequent one, only impacts performance in a small way, and thus, we opted for the simpler approach.

\begin{table}[t]
\caption{Dataset statistics. Note that ``EN'' refers to English labels in Wikidata, but in our setup it actually means Latin script. That is why the source alphabet size exceeds 100 for all the four Wikidata datasets. As expected, the ``EN-RU'' dataset has the largest Latin alphabet size (source alphabet size).}
\label{tbl:data-stats}
\begin{center}
\bgroup
\def\arraystretch{1.0}
\begin{tabular}{|C{3cm}|C{1.65cm}|C{1.65cm}|C{1.4cm}|C{1.4cm}|C{1.65cm}|C{1.65cm}|}
\hline
\textbf{Dataset} & \textbf{Total Size} & \textbf{Training Set Size} & \textbf{Avg. Source Length} & \textbf{Avg. Target Length} & \textbf{Source Alphabet Size} & \textbf{Target Alphabet Size} \\ \hline
WD-EN-RU         & 164,640 & 105,371        & 7.0                 & 6.6                & 188     & 62       \\ \hline
WD-EN-KA         & 98,820  & 63,246        & 7.0                 & 4.8                & 170     & 105        \\ \hline
WD-EN-AR         & 74,973 & 41,584          & 6.7                 & 5.9                & 145     & 67        \\ \hline
WD-EN-HE         & 50,039 & 32,036         & 6.7                 & 5.6                & 135     & 57        \\ \hline
Rosca and B., 2016            & 15,898 & 12,877         & 6.0                 & 6.8                & 48      & 39        \\ \hline
CMUdict          & 126,191 & 113,438        & 7.5                 & 6.3                & 27      & 39        \\ \hline
\end{tabular}
\egroup
\end{center}
\end{table}

We used a typical 64/16/20 cross-validation split of the single token data. We used 64\% of the data for training, 16\% for a development evaluation set to pick hyperparameters, and a 20\% held-out test set for metric reporting in Section~\ref{sec:results}. Our curated datasets are being released publicly in two forms. We are releasing the original name phrases where each line contains two tab-separated columns with the English name phrase in the first and the target script name phrase in the second column. Additionally, we are releasing the particular 64/16/20 splits of the aligned single token data in order that future researchers can replicate and benchmark against our results. Table \ref{tbl:data-stats} provides more information about the aligned, single token versions of each dataset.

For evaluating performance on the Arabic to English dataset from~\cite{rosca2016sequence}, we used the same train, dev, and test splits used in the paper. For evaluating CMUdict performance, we used the same 90\% train, 10\% split used in \cite{novak2012wfst}. We used these datasets without making any changes. 

\subsection{Libraries}

We evaluate the previously discussed approaches using the following libraries:
\begin{itemize}
    \item \texttt{Phonetisaurus} library\footnote{\url{https://github.com/AdolfVonKleist/Phonetisaurus}}~\cite{novak2012wfst}, the traditional WFST approach to sequence to sequence transduction.
    \item \texttt{seq2seq} library\footnote{\url{https://github.com/tensorflow/nmt}}, the encoder-decoder recurrent neural network approach~\cite{luong17} as implemented in the TensorFlow~\cite{tensorflow2015-whitepaper} machine learning platform.
    \item \texttt{tensor2tensor} library\footnote{\url{https://github.com/tensorflow/tensor2tensor}}, the reference implementation of the Transformer approach to neural sequence to sequence transduction tasks~\cite{vaswani2017attention}, which is implemented on top of TensorFlow.
\end{itemize}

For the experiments using Phonetisaurus, we used the default configuration with an 8-order MITLM~\cite{hsu2008iterative} language model. For the many-to-many alignment, we disallowed $\epsilon$-transitions on the source side and allowed them on the target side (with the \texttt{--seq2\_del} option). For the encoder-decoder experiments, we simply adapted the Seq2Seq tutorial~\cite{luong17} to use the library for our character-level transliteration task. We used an LSTM cell-type, 2 hidden layers, 128 units per layer, a dropout probability of 0.2, and default `luong' attention mechanism. Lastly, for the Tensor2Tensor experiments, we used the default  configuration but tested with both 64 units per layer and 128 units per layer. Using 128 units per layer improved WER for every language by $\approx$~2-3 points, and thus we only report results with 128 units. All of the scripts used in these experiments along with the mined Wikidata datasets are publicly available\footnote{\url{https://github.com/steveash/NETransliteration-COLING2018}}.

\subsection{Evaluation Metric}

Since many names may have multiple correct transliterations, in most experiments, we report the Word Error Rate (WER) metric, which is based on the proportion of name tokens that were transliterated and exactly match the expected target script from the test set (ground truth). The lower the WER score, the better.
We report WER scores as $1$-best, $2$-best, and $3$-best, which reflects when the correct transliteration occurs in the top spot, or in one of the top 2 spots, or in one of the top 3 spots. Since we do not penalize bad transliterations, 2-best is always at least as accurate as 1-best, and 3-best is always at least as accurate as 1-best and 2-best. We feel it is important to report in this way, because in many cases, such as information retrieval, it is feasible to generate multiple possible transliterations and use all of them at search time to improve recall. In other cases, such as text-to-speech, the top-1 result is the only result that matters, because the system only has a single opportunity to provide the correct transliteration.

\section{Experimental Results}\label{sec:results}




Table~\ref{tbl:different-methods} summarizes the results of each method on different datasets. The recent Tensor2Tensor Transformer architecture outperforms the WFST approach and the Seq2Seq approach on every language. However, it is worth noting that the training time using Tensor2Tensor is between 5-8 hours using an AWS p3.2xlarge\footnote{\url{https://aws.amazon.com/ec2/instance-types/p3/}} instance type, which has a Tesla V100 GPU. The Phonetisaurus training time only takes a couple of minutes on a typical dual-core Intel Core-i7 personal laptop. We did not expect the WER difference between T2T and Phonetisaurus to be that significant. Our hypothesis was that given the small number of characters in every language and small average input size, the n-gram based models would be hard to beat. NMT approaches have a clear advantage in handling long term dependencies, but we expected that the small input size of our named entity tokens implied few important long term dependencies. The CMUdict dataset is the only case in which Phonetisaurus (slightly) outperforms the neural Transformer approach. One explanation is that the CMUdict dataset is not that challenging, as can be seen by the difference in WER compared to all other datasets. 

\begin{table}[t]
\caption{Word Error Rate (WER) using different methods for named entity transliteration. ``WD'' refers to our produced Wikidata datasets. ``T2T'' refers to the Tensor2Tensor system.}
\label{tbl:different-methods}
\begin{center}
\bgroup
\def\arraystretch{1.0}
\begin{tabular}{|C{3.5cm}|C{3.5cm}|C{3cm}|C{1.2cm}|C{1.2cm}|C{1.2cm}|}
\hline
\textbf{Task} & \textbf{Dataset} & \textbf{Method} & \textbf{1-best WER} & \textbf{2-best WER} & \textbf{3-best WER} \\ \hline
\multirow{3}{*}{English $\rightarrow$ Arabic} & \multirow{3}{*}{WD-EN-AR}          & T2T    & \textbf{0.45}       & \textbf{0.30}       & \textbf{0.24}       \\ 
&          & Seq2Seq         & 0.53                & 0.41   & 0.36                    \\ 
&         & Phonetisaurus   & 0.51                & 0.37                & 0.30                \\ \hline

\multirow{3}{*}{English $\rightarrow$ Hebrew} &  \multirow{3}{*}{WD-EN-HE}         & T2T    & \textbf{0.44}       & \textbf{0.27}       & \textbf{0.22} 
\\ 
&          & Seq2Seq         & 0.49                & 0.36       & 0.31                    \\ 
&          & Phonetisaurus   & 0.49                & 0.32                & 0.26                \\ \hline

\multirow{3}{*}{English $\rightarrow$ Katakana} & \multirow{3}{*}{WD-EN-KA}        & T2T    & \textbf{0.51}       & \textbf{0.35}       & \textbf{0.29}       \\ 
&          & Seq2Seq         & 0.60                & 0.48                 & 0.43                    \\ 
&          & Phonetisaurus   & 0.57                & 0.43                & 0.36                \\ \hline

\multirow{3}{*}{English $\rightarrow$ Russian} & \multirow{3}{*}{WD-EN-RU}         & T2T    & \textbf{0.35}       & \textbf{0.22}       & \textbf{0.17}       \\ 
&        & Seq2Seq         & 0.40                & 0.29        & 0.25                     \\ 
&          & Phonetisaurus   & 0.38                & 0.24                & 0.19                \\ \hline

\multirow{3}{*}{Arabic $\rightarrow$ English} &  \multirow{3}{*}{Rosca and B., 2016}           & T2T   & \textbf{0.75}       & \textbf{0.63}       & \textbf{0.57}       \\ 
&             & Seq2Seq         & 0.81                &  0.71     &  0.67                   \\ 
&             & Phonetisaurus   & 0.75                & 0.65                & 0.59                \\ \hline

\multirow{3}{*}{English $\rightarrow$ ARPAbet} & \multirow{3}{*}{CMUdict}         & T2T    & 0.29                & 0.16                & 0.11                \\ 
&          & Seq2Seq         & 0.29                &  0.18       & 0.14                    \\ 
&          & Phonetisaurus   & \textbf{0.27}       & \textbf{0.14}       & \textbf{0.10}       \\ \hline
\end{tabular}
\egroup
\end{center}
\end{table}

\begin{figure}[t]
    \centering
    \includegraphics[width=0.6\textwidth]{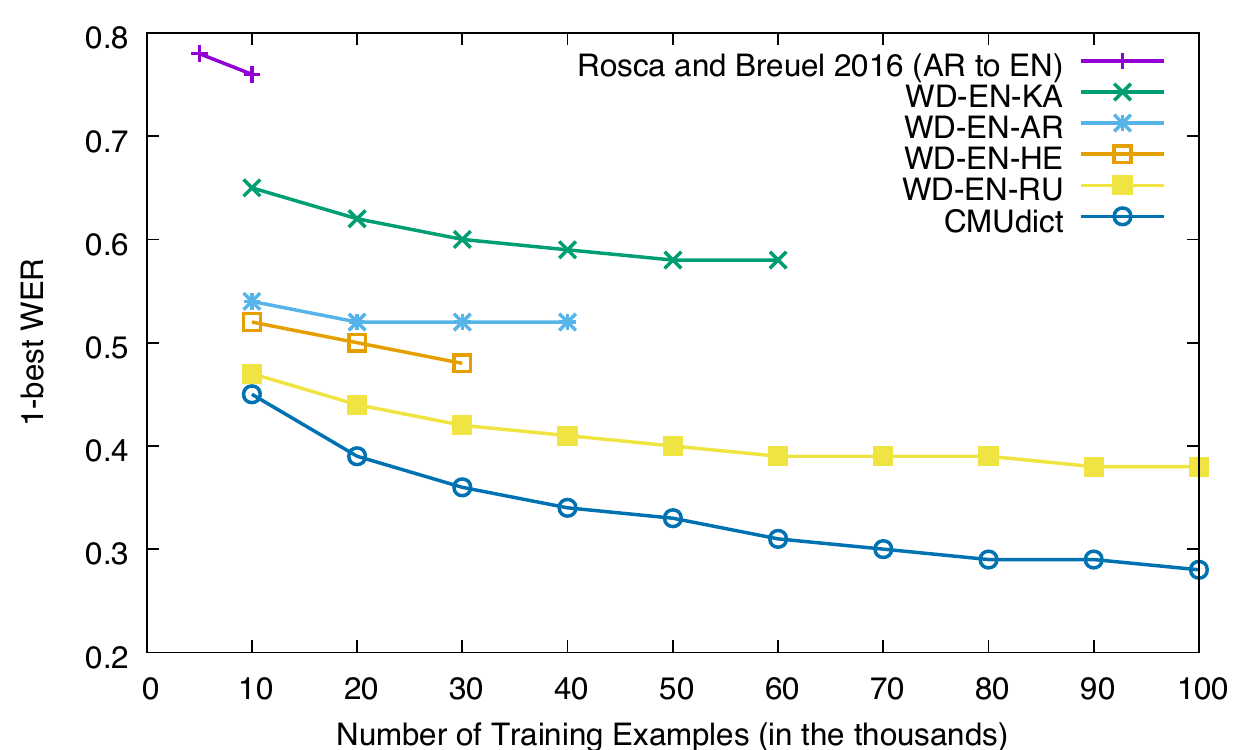}
    \caption{Phonetisaurus: Impact of the number of training examples on WER.}
    \label{fig:curves}
\end{figure}

Another observation is the WER gap between the languages. Among the English to $X$ tasks, Katakana has the worse WER, followed by Arabic that is only slightly behind Hebrew. Given that Arabic and Hebrew belong to the same language family, it is not surprising that their WER is similar. Russian has the best WER by a large margin. It also has the largest training data and it is the closest language to English among the four evaluated. We were interested to learn the impact of training size on the error rate. Figure~\ref{fig:errors} shows learning curves for various datasets. We used Phonetisaurus since it was the fastest to train. One interesting observation is that training size is not a big factor. When the number of training examples is equal across all languages, Russian still has the best WER by a large margin and Katakana the worst by a large margin. Another interesting observation is that we reach close to optimal performance with about 50\% of the training data. For example, WER on WD-EN-RU (English to Russian) is 0.40 and 0.38 after training on 50K and 100K examples, respectively. CMUdict has the steepest reduction in WER as the number of training examples increases, with WER of 0.33 and 0.28 after training on 50K and 100K examples, respectively.

The seq2seq system consistently performed the worst. Based on design guidance in Neural Machine Translation literature~\cite{sutskever2014sequence}, we experimented with feeding the source string in reverse order as this showed a significant improvement in encoder-decoder approaches. In our experiments, this did not meaningfully impact the WER. The resulting score was either identical or within one point for each of the language pairs that we tested.  

\subsection{Removing tokens that occur only once}\label{sec:onlyOnceResults}

One design question that we set out to answer is how to be handle unique name token pairs in dataset curation (e.g., the `EN-KA' transliteration pair [``escalada'', ``\begin{CJK}{UTF8}{min}エスカラーダ\end{CJK}''] occurs only once in the data we collected from Wikidata). Name tokens, like other language tokens, follow a Zipfian distribution and thus there are likely to be many tokens that appear infrequently. On the one hand, tokens that occur only once may be erroneous, and it may be better to exclude them from the training and test sets. Taking the English to Hebrew dataset as an example, 77\% of the name tokens only occur once. Our hypothesis was that given this distribution of token frequency, excluding them would likely remove more useful information than noise. Table~\ref{tbl:infrequent-tokens} illustrates the impact of excluding single occurrence name tokens from both train and test against a baseline of training and testing on everything. All of these results use the Transformer neural approach with 128 units per layer and 2 hidden layers.

Comparing column 1 (baseline) and column 2 in Table~\ref{tbl:infrequent-tokens} shows that excluding the infrequent tokens from test, results in a surprisingly high reduction in WER. If we also filter the tokens from train (as shown in column 4), we see an increase in WER. The worse WER is achieved when filtering the train set but testing on the full test set. In English to Hebrew (WD-EN-HE), filtering the train set but testing on the full set increases the WER by 7.0 points over the baseline. 
Industrial transliteration systems will contain dictionary look-ups for frequent named entity transliterations and employ automated methods, as evaluated here, to handle out-of-vocabulary cases. Thus, it stands to reason that a fair evaluation of automated methods should include infrequent names even in the presence of some noise. The results presented in other sections include unique tokens at both train and test time.

An interesting question is why infrequent names in our data happen to be the hardest instances. One possible factor is that these names are less known to the public so there is no spelling convention to follow, leading to higher ambiguity. Wikidata labels are updated by many people from all over the world, hence name frequency can be thought as a proxy for annotator agreement. We also found that infrequent names in our data are longer on average, with a strong negative correlation between token frequency and token length. For example, the average length of all single occurrence English name tokens based on all of our test datasets is 7.1 characters, while the average length of all other English name tokens (at least two occurrences) is 6.4. Word-based evaluation metrics like we use here might be biased towards shorter names. Some prior works have used character-based metrics for transliteration, such as length-normalized edit distance~\cite{irvine2010transliterating,noeman2010language}, which might be a good alternative (depending on the application).  

\subsection{English as source or target language}\label{sec:englishsource}

\begin{table}[t]
\caption{T2T 1-Best Word Error Rate (WER) when filtering single occurrence name tokens from training and/or testing.}
\label{tbl:infrequent-tokens}
\begin{center}
\bgroup
\def\arraystretch{1.0}
\begin{tabular}{|C{2.2cm}|C{2cm}|C{2cm}|C{2cm}|C{2cm}|C{2cm}|}
\hline
\textbf{Dataset} & 
    \textbf{Full Train, Full Test (Baseline)} &
    \textbf{Full Train, Filtered Test} &
    \textbf{Filtered Train, Full Test} & 
    \textbf{Filtered Train, Filtered Test} \\
\hline
WD-EN-AR  & 0.45 & 0.32 & 0.50 & 0.35  \\
\hline
WD-EN-HE  & 0.43 & 0.31 & 0.50 & 0.35  \\
\hline
WD-EN-RU  & 0.35 & 0.24 & 0.39 & 0.26   \\
\hline
WD-EN-KA  & 0.50 & 0.36 & 0.56 & 0.41 \\
\hline
\end{tabular}
\egroup
\end{center}
\end{table}

\begin{table}[t]
\caption{T2T Word Error Rate (WER) when modeling as English to $X$ versus $X$ to English.}
\label{tbl:english-to-x}
\begin{center}
\bgroup
\def\arraystretch{1.0}
\begin{tabular}{|C{3.0cm}|C{4cm}|C{1.3cm}|C{1.3cm}|C{1.3cm}|}
\hline
\textbf{Dataset} & \textbf{Task} & \textbf{1-best WER} & \textbf{2-best WER} & \textbf{3-best WER} \\ \hline
WD-EN-AR         & English $\rightarrow$ Arabic     & \textbf{0.45}       & \textbf{0.30}       & \textbf{0.24}       \\ 
WD-EN-AR         & Arabic $\rightarrow$ English     & 0.75                & 0.64                & 0.58                \\ \hline
WD-EN-HE         & English $\rightarrow$ Hebrew     & \textbf{0.44}       & \textbf{0.27}       & \textbf{0.22}       \\ 
WD-EN-HE         & Hebrew $\rightarrow$ English     & 0.77                & 0.65                & 0.59                \\ \hline
WD-EN-KA         & English $\rightarrow$ Katakana   & \textbf{0.51}       & \textbf{0.35}       & \textbf{0.29}       \\ 
WD-EN-KA         & Katakana $\rightarrow$ English   & 0.70                & 0.57                & 0.50                \\ \hline
WD-EN-RU         & English $\rightarrow$ Russian    & \textbf{0.35}       & \textbf{0.22}       & \textbf{0.17}       \\ 
WD-EN-RU         & Russian $\rightarrow$ English    & 0.47                & 0.35                & 0.30                \\ \hline
\end{tabular}
\egroup
\end{center}
\end{table}

We performed experiments comparing the performance of modeling the transliteration problem as English~$\rightarrow X$ versus $X \rightarrow$~English. Our hypothesis was that learning the transliteration with English as the source language was going to result in a lower word error rate due to the loss of information when going to languages like Hebrew or Arabic. Table~\ref{tbl:english-to-x} reports the differences in these two approaches. In every case using English as the source language results in much better word error rate, but the impact varies depending on the language. Hebrew and Arabic are impacted the most, and Russian is the least penalized. Previous works that focus on back-transliterating (recovering names of English origin) report similar findings~\cite{irvine2010transliterating}. In the error analysis section we provide some insights on why the reverse task is harder. 

\subsection{Error Analysis}
Given the nature of Wikidata, our extracted datasets contain a diverse set of names from different origins, including many foreign names with different linguistic conventions. Names can be pronounced differently depending on origin and context can play a role as well. This explains why only outputting the top transliteration is usually not enough, as our results show. One example showing how hard the task can be is the Irish name ``Domhnall'', pronounced as DONAL, which none of the models transliterated correctly. The Hebrew, Arabic, and Russian models all included an `m' in each of their 3-best outputs. 

\begin{figure}
    \centering
    \includegraphics[width=0.8\textwidth]{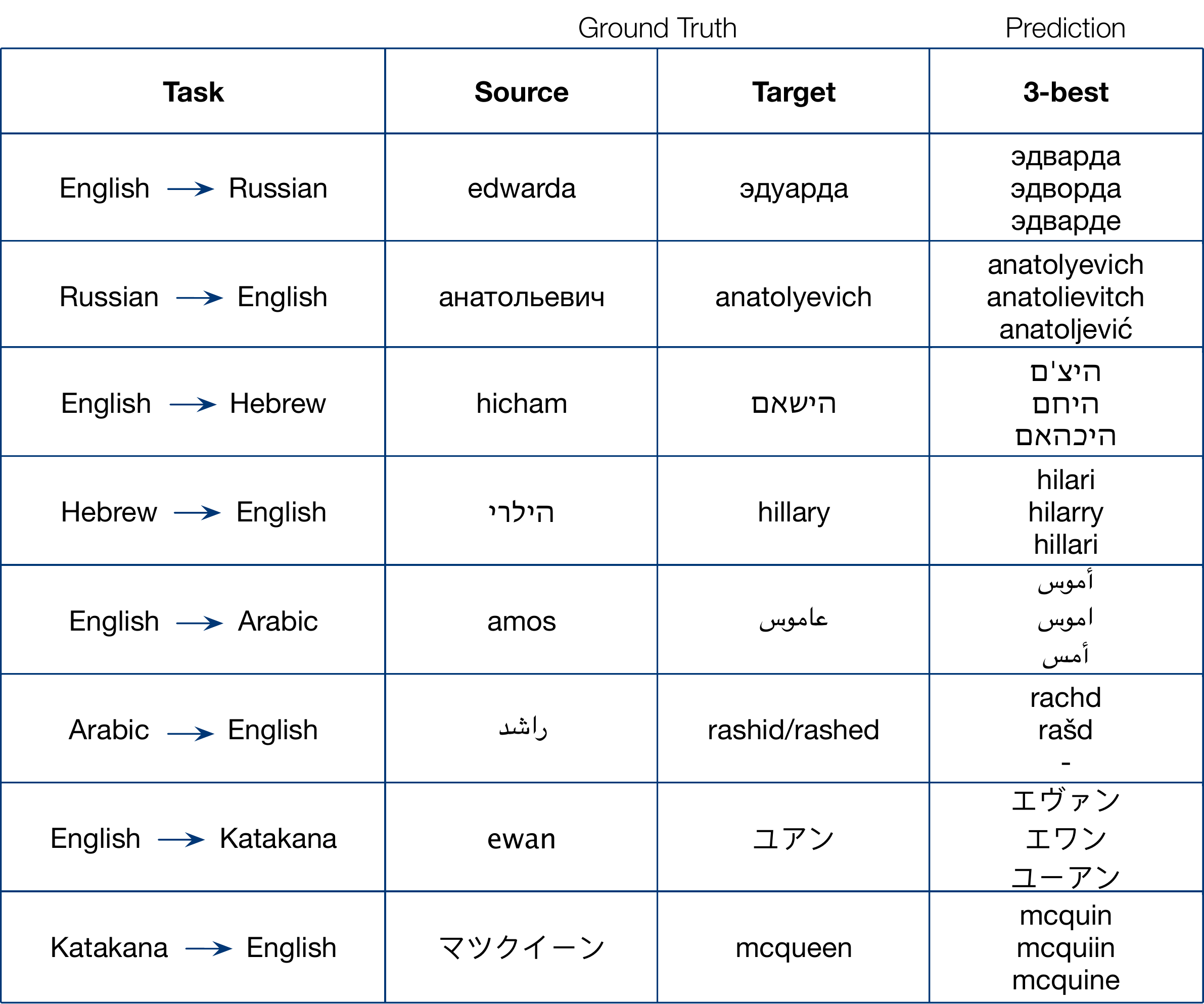}
    \caption{Model errors across different languages}
    \label{fig:errors}
\end{figure}

Figure~\ref{fig:errors} shows error examples in the various tasks. Vowel confusion is a common error.  As expected, the Hebrew and Arabic models are affected by it the most. The problem becomes even worse when English is the target language. Hebrew and Arabic names in Wikidata, like in most of the web, do not include diacritical marks. This means that often English names with vowels are written in Arabic and Hebrew without the vowels. For example, ``barak'' is written as ``brk'' (in Hebrew letters) in modern Hebrew. When transliterating to English, the model often needs to recover the missing vowels. One of the figure examples is the name ``Rashid''/``Rashed'' that was transliterated incorrectly since the model failed to recover the letter `i'/`e' that is not included in the original Arabic name. Vowel confusion is also a problem in Japanese, as the ``ewan'' example in  the figure shows (the name Ewan MacGregor is quite known in Japan). The third best transliteration is ``\begin{CJK}{UTF8}{min}ユーアン\end{CJK}'', which differs from the correct transliteration only by the Katakana long vowel ``\begin{CJK}{UTF8}{min}ー\end{CJK}''. There seem to be different vowel variations in Katakana, such as long vs. short vowels. We found it often happens in the word final position. Having a special treatment to handle such cases could significantly boost accuracy. Some errors are more language specific. For example, in English to Russian the model often fails on the letter `w' that can be ambiguous since the sound does not exist in Russian. The figure shows how the model made the same error in all three transliterations of the name ``edwarda'', replacing `w' with the Russian letter `B' ('v' sound), which is commonly used as a replacement. 

Another interesting observation is that the models also fail on common names. This happens in all languages but more frequently occurs when English is the target language. For example, the Hebrew to English model transliterated ``hillary'' from Hebrew to English as ``hilari'', ``hilarry'', and ``hillari''. We believe this is a result of our problem setup. Since all our test names are unseen in training and every name only occurs once without making use of frequency info, the language model cannot learn that ``hillary'' is more common than ``hilarry'', for example, unless it is a sub-token of a longer name in training. Incorporating name frequency or large character-based n-gram language models learned from a large corpus (as in~\cite{al2002translating,irvine2010transliterating}) could possibly help such cases.

\section{Future Directions}
We will continue exploring best practices for multilingual transliteration. We are planning to evaluate design criteria for doing transliteration with CJK languages, where alignment is a more difficult problem. We are also interested to understand better the relationship between token frequency and transliteration quality. Resources like Wikidata are extremely useful for curating large multilingual datasets, but some thought should be given to how we can reduce bias and noise with minimal effort, aiming to achieve quality that is comparable with manually annotated datasets. We believe that token frequency can be used as a proxy for annotator agreement and are planning to study this further. Finally, we are interested to study how well character-based NMT systems such as~\cite{ling2015character} perform on named entity transliteration.

\section{Conclusions}
We described a number of design considerations that one must address when building a robust, multilingual named entity transliteration system. We empirically demonstrated that the Tensor2Tensor neural Transformer method produces consistently strong results against the LSTM-based encoder-decoder neural method and WFST-based traditional method. We described a number of challenges and design choices when building bilingual dictionaries from mined knowledge bases, such as Wikidata. Our Wikidata curated datasets, including both the name phrase bilingual data and the aligned, single token datasets are being released for further experimentation and benchmarking.

\bibliography{references}

\begin{thebibliography}{}

\bibitem[\protect\citename{Abadi \bgroup et al.\egroup
  }2015]{tensorflow2015-whitepaper}
Mart\'{\i}n Abadi, Ashish Agarwal, Paul Barham, Eugene Brevdo, Zhifeng Chen,
  Craig Citro, Greg~S. Corrado, Andy Davis, Jeffrey Dean, Matthieu Devin,
  Sanjay Ghemawat, Ian Goodfellow, Andrew Harp, Geoffrey Irving, Michael Isard,
  Yangqing Jia, Rafal Jozefowicz, Lukasz Kaiser, Manjunath Kudlur, Josh
  Levenberg, Dandelion Man\'{e}, Rajat Monga, Sherry Moore, Derek Murray, Chris
  Olah, Mike Schuster, Jonathon Shlens, Benoit Steiner, Ilya Sutskever, Kunal
  Talwar, Paul Tucker, Vincent Vanhoucke, Vijay Vasudevan, Fernanda Vi\'{e}gas,
  Oriol Vinyals, Pete Warden, Martin Wattenberg, Martin Wicke, Yuan Yu, and
  Xiaoqiang Zheng.
\newblock 2015.
\newblock {TensorFlow}: Large-scale machine learning on heterogeneous systems.
\newblock Software available from tensorflow.org.

\bibitem[\protect\citename{Al-Onaizan and Knight}2002]{al2002translating}
Yaser Al-Onaizan and Kevin Knight.
\newblock 2002.
\newblock Translating named entities using monolingual and bilingual resources.
\newblock In {\em Proceedings of the 40th Annual Meeting on Association for
  Computational Linguistics}, pages 400--408. Association for Computational
  Linguistics.

\bibitem[\protect\citename{Bahdanau \bgroup et al.\egroup }2014]{bahdanauCB14}
Dzmitry Bahdanau, Kyunghyun Cho, and Yoshua Bengio.
\newblock 2014.
\newblock Neural machine translation by jointly learning to align and
  translate.
\newblock {\em CoRR}, abs/1409.0473.

\bibitem[\protect\citename{Bisani and Ney}2008]{bisani2008joint}
Maximilian Bisani and Hermann Ney.
\newblock 2008.
\newblock Joint-sequence models for grapheme-to-phoneme conversion.
\newblock {\em Speech communication}, 50(5):434--451.

\bibitem[\protect\citename{Ekbal \bgroup et al.\egroup
  }2006]{ekbal2006modified}
Asif Ekbal, Sudip~Kumar Naskar, and Sivaji Bandyopadhyay.
\newblock 2006.
\newblock A modified joint source-channel model for transliteration.
\newblock In {\em Proceedings of the COLING/ACL on Main conference poster
  sessions}, pages 191--198. Association for Computational Linguistics.

\bibitem[\protect\citename{Fushimi \bgroup et al.\egroup
  }1999]{fushimi1999consistency}
Takao Fushimi, Matsuo Ijuin, Karalyn Patterson, and Itaru~F Tatsumi.
\newblock 1999.
\newblock Consistency, frequency, and lexicality effects in naming japanese
  kanji.
\newblock {\em Journal of Experimental Psychology: Human Perception and
  Performance}, 25(2):382.

\bibitem[\protect\citename{Haizhou \bgroup et al.\egroup
  }2004]{haizhou2004joint}
Li~Haizhou, Zhang Min, and Su~Jian.
\newblock 2004.
\newblock A joint source-channel model for machine transliteration.
\newblock In {\em Proceedings of the 42nd Annual Meeting on association for
  Computational Linguistics}, page 159. Association for Computational
  Linguistics.

\bibitem[\protect\citename{Hsu and Glass}2008]{hsu2008iterative}
Bo-June Hsu and James Glass.
\newblock 2008.
\newblock Iterative language model estimation: efficient data structure \&
  algorithms.
\newblock In {\em Ninth Annual Conference of the International Speech
  Communication Association}.

\bibitem[\protect\citename{Irvine \bgroup et al.\egroup
  }2010]{irvine2010transliterating}
Ann Irvine, Chris Callison-Burch, and Alexandre Klementiev.
\newblock 2010.
\newblock Transliterating from all languages.
\newblock In {\em Proceedings of the Conference of the Association for Machine
  Translation in the Americas (AMTA)}, pages 100--110.

\bibitem[\protect\citename{Knight and Graehl}1998]{knight1998machine}
Kevin Knight and Jonathan Graehl.
\newblock 1998.
\newblock Machine transliteration.
\newblock {\em Computational Linguistics}, 24(4):599--612.

\bibitem[\protect\citename{Ling \bgroup et al.\egroup }2015]{ling2015character}
Wang Ling, Isabel Trancoso, Chris Dyer, and Alan~W Black.
\newblock 2015.
\newblock Character-based neural machine translation.
\newblock {\em arXiv preprint arXiv:1511.04586}.

\bibitem[\protect\citename{Luong \bgroup et al.\egroup }2017]{luong17}
Minh{-}Thang Luong, Eugene Brevdo, and Rui Zhao.
\newblock 2017.
\newblock Neural machine translation (seq2seq) tutorial.
\newblock {\em https://github.com/tensorflow/nmt}.

\bibitem[\protect\citename{Noeman and Madkour}2010]{noeman2010language}
Sara Noeman and Amgad Madkour.
\newblock 2010.
\newblock Language independent transliteration mining system using finite state
  automata framework.
\newblock In {\em Proceedings of the 2010 Named Entities Workshop}, pages
  57--61. Association for Computational Linguistics.

\bibitem[\protect\citename{Novak \bgroup et al.\egroup }2012]{novak2012wfst}
Josef~R Novak, Nobuaki Minematsu, and Keikichi Hirose.
\newblock 2012.
\newblock Wfst-based grapheme-to-phoneme conversion: Open source tools for
  alignment, model-building and decoding.
\newblock In {\em Proceedings of the 10th International Workshop on Finite
  State Methods and Natural Language Processing}, pages 45--49.

\bibitem[\protect\citename{Pasternack and Roth}2009]{pasternack2009learning}
Jeff Pasternack and Dan Roth.
\newblock 2009.
\newblock Learning better transliterations.
\newblock In {\em Proceedings of the 18th ACM conference on Information and
  knowledge management}, pages 177--186. ACM.

\bibitem[\protect\citename{Rao \bgroup et al.\egroup }2015]{rao2015grapheme}
Kanishka Rao, Fuchun Peng, Ha{\c{s}}im Sak, and Fran{\c{c}}oise Beaufays.
\newblock 2015.
\newblock Grapheme-to-phoneme conversion using long short-term memory recurrent
  neural networks.
\newblock In {\em Acoustics, Speech and Signal Processing (ICASSP), 2015 IEEE
  International Conference on}, pages 4225--4229. IEEE.

\bibitem[\protect\citename{Rosca and Breuel}2016]{rosca2016sequence}
Mihaela Rosca and Thomas Breuel.
\newblock 2016.
\newblock Sequence-to-sequence neural network models for transliteration.
\newblock {\em arXiv preprint arXiv:1610.09565}.

\bibitem[\protect\citename{Stalls and Knight}1998]{stalls1998translating}
Bonnie~Glover Stalls and Kevin Knight.
\newblock 1998.
\newblock Translating names and technical terms in arabic text.
\newblock In {\em Proceedings of the Workshop on Computational Approaches to
  Semitic Languages}, pages 34--41. Association for Computational Linguistics.

\bibitem[\protect\citename{Sutskever \bgroup et al.\egroup
  }2014]{sutskever2014sequence}
Ilya Sutskever, Oriol Vinyals, and Quoc~V Le.
\newblock 2014.
\newblock Sequence to sequence learning with neural networks.
\newblock In {\em Advances in neural information processing systems}, pages
  3104--3112.

\bibitem[\protect\citename{Thu \bgroup et al.\egroup }2016]{thu2016comparison}
Ye~Kyaw Thu, Win~Pa Pa, Yoshinori Sagisaka, and Naoto Iwahashi.
\newblock 2016.
\newblock Comparison of grapheme-to-phoneme conversion methods on a myanmar
  pronunciation dictionary.
\newblock In {\em Proceedings of the 6th Workshop on South and Southeast Asian
  Natural Language Processing (WSSANLP2016)}, pages 11--22.

\bibitem[\protect\citename{Vaswani \bgroup et al.\egroup
  }2017]{vaswani2017attention}
Ashish Vaswani, Noam Shazeer, Niki Parmar, Jakob Uszkoreit, Llion Jones,
  Aidan~N Gomez, {\L}ukasz Kaiser, and Illia Polosukhin.
\newblock 2017.
\newblock Attention is all you need.
\newblock In {\em Advances in Neural Information Processing Systems}, pages
  6000--6010.

\bibitem[\protect\citename{Weide}2014]{weide1998cmu}
R~Weide.
\newblock 2014.
\newblock The {CMU} pronunciation dictionary, release 0.7b.

\bibitem[\protect\citename{Yan and Nivre}2016]{yan2016applying}
Shao Yan and Joakim Nivre.
\newblock 2016.
\newblock Applying neural networks to english-chinese named entity
  transliteration.
\newblock In {\em Sixth Named Entity Workshop, joint with 54th ACL}.

\bibitem[\protect\citename{Yao and Zweig}2015]{yao2015sequence}
Kaisheng Yao and Geoffrey Zweig.
\newblock 2015.
\newblock Sequence-to-sequence neural net models for grapheme-to-phoneme
  conversion.
\newblock {\em arXiv preprint arXiv:1506.00196}.

\end{thebibliography}
\bibliographystyle{acl}

\end{document}